%% file: main.tex
\documentclass[runningheads]{llncs}
\usepackage{graphicx}
\usepackage{comment}
\usepackage{amsmath,amssymb} 
\usepackage{color}

\usepackage{breakcites}
\usepackage{mwe}
\usepackage{pifont}
\newcommand{\cmark}{\ding{51}}%
\newcommand{\xmark}{\ding{55}}%

\usepackage{algorithm}
\usepackage{amsmath}
\usepackage[noend]{algpseudocode}
\usepackage{hhline}
\usepackage{multirow}
\usepackage{hyperref}

\newcommand\sbullet[1][.75]{\mathbin{\vcenter{\hbox{\scalebox{#1}{$\bullet$}}}}}

\usepackage{afterpage}

\DeclareRobustCommand{\name}{FAD\@\xspace}
\DeclareRobustCommand{\rep}{RepShare\@\xspace}
\DeclareRobustCommand{\eg}{e.g.\@\xspace}
\DeclareRobustCommand{\ie}{i.e.\@\xspace}

\DeclareRobustCommand{\etal}{et al.\@\xspace}
\DeclareRobustCommand{\vs}{vs.\@\xspace}

\begin{document}
\pagestyle{headings}
\mainmatter
\def\ECCVSubNumber{3266}  

\title{Representation Sharing for \\ Fast  Object Detector Search and Beyond}

\titlerunning{Representation Sharing for Fast  Object Detector Search and Beyond}
\author{Yujie Zhong \and 
Zelu Deng \and
Sheng Guo \and
Matthew R. Scott \and 
Weilin Huang\thanks{Corresponding author: whuang@malong.com}}
\authorrunning{Y. Zhong et al.}
\institute{Malong LLC\\
{\tt\small \{jaszhong, zeldeng, sheng, mscott, whuang\}@malongtech.com}
}

\maketitle

\input{abstract.tex}

\input{intro.tex}

\input{related.tex}

\input{searchSpace.tex}

\input{sharing.tex}

\input{exp.tex}

\input{conclusion.tex}

\clearpage

\bibliographystyle{splncs04}
\bibliography{mybib}

\clearpage
\input{app.tex}

\end{document}

%% file: abstract.tex
\vspace{-4mm}

\begin{abstract}

Region  Proposal  Network (RPN) provides strong support for handling the scale variation of objects in two-stage object detection. For one-stage detectors which do not have RPN, it is more demanding to have powerful sub-networks capable  of directly  capturing  objects of unknown sizes.  
To enhance such capability, 
we propose an extremely efficient neural architecture search method, named Fast And Diverse (\name),
to better explore the optimal configuration of receptive fields and convolution types in the sub-networks for one-stage detectors.
\name consists of a designed search space and an efficient architecture search algorithm. 
The search space contains a rich set of diverse transformations designed specifically for object detection.
To cope with the designed search space, a novel search algorithm termed Representation Sharing (\rep) is proposed to effectively identify the best combinations of the defined transformations.
In our experiments,  \name obtains prominent improvements on two types of one-stage detectors with various backbones. %
In particular, our \name detector  achieves 46.4 AP on MS-COCO (under single-scale testing), outperforming the state-of-the-art detectors, including the most recent NAS-based detectors, Auto-FPN~\cite{xu2019auto} (searched for 16 GPU-days) and NAS-FCOS~\cite{wang2019fcos} (28 GPU-days), while significantly reduces the search cost to 0.6 GPU-days.
Beyond object detection, we further demonstrate the generality of \name on the more challenging instance segmentation, and expect it to benefit more tasks.
The code is available at \href{https://github.com/MalongTech/research-fad}{https://github.com/MalongTech/research-fad}.

\end{abstract}

%% file: intro.tex
\section{Introduction}

Object detection is a fundamental task in computer vision \cite{ren2015faster,liu2016ssd,lin2017feature,redmon2018yolov3,lin2017focal,liu2018receptive,li2019scale,tian2019fcos,zhang2019freeanchor}, 
but it remains challenging due to the large variation in object scales. 
To handle the scale variation, a straightforward method is to utilize multi-scale image inputs~\cite{singh2018analysis,singh2018sniper}, which usually lacks efficiency. A line of more efficient methods is to tackle the scale variation on the intermediate features~\cite{liu2016ssd,lin2017feature}. For example, Feature
Pyramid Networks (FPN)~\cite{lin2017feature} is a representative work that implements the detection of objects with different scales in multiple levels of feature pyramids.
On the other hand, recent works also attempt to improve the detectors from the perspective of receptive fields (RFs)~\cite{liu2018receptive,li2019scale}. They enhance the scale-awareness of the detectors by having multi-branch transformations with different combinations of kernel sizes and/or dilation rates. Then the features of different RFs are aggregated to enrich the information of different scales at each spatial location.

An  object detector often has a backbone network followed by the detection-specific sub-networks (\ie heads), which play an important role in object detection. The sub-networks compute the deep features which are used to directly predict the object category, localization and size.
Unlike two-stage detectors in which the sub-networks operate on the fixed-size feature maps computed from each object proposal, generated by a region proposal network with ROI-pooling~\cite{ren2015faster}, the sub-networks in one-stage detectors should be capable of `looking for' objects of arbitrary sizes directly.
It becomes more challenging for an anchor-free detector. Because the multi-scale anchor boxes can be considered as a way to explicitly handle various sizes and shapes of objects, 
whereas an anchor-free detector only predicts a single object at each spatial location, without any prior information about the object size.
Therefore, for one-stage detectors, especially the anchor-free ones, the capability of the sub-networks for capturing the objects with large scale variation becomes the key.
In this work, we aim to enhance the power of the sub-networks in one-stage detectors, by searching for the optimal combination of the RFs and convolutions in a learning-based manner.

\input{tables/gpudays}

Neural Architecture Search (NAS) has gained increasing attention. 
It transfers the task of neural networks design from a heuristics-guided process to an optimization problem. 
Recently, it has been shown that NAS can achieve prominent results on object detection~\cite{ghiasi2019fpn,chen2019detnas,xu2019auto,peng2019efficient,yao2019sm,wang2019fcos}.
In most of the work, the operations in the search space are directly extended from those used for image classification~\cite{zoph2016neural,liu2018darts} with limited variation on dilation rates. Therefore, their search spaces with respect to \emph{transformations} are relatively limited, as listed in Table~\ref{tab:gpudays}.
Apart from the combination of RFs, we also investigate the importance of the diversity of the transformations in NAS search space for object detection. 
However, searching through such a large number of candidate transformations can be computationally expensive,
especially for the RL-based~\cite{zoph2016neural,pham2018efficient} and EA-based~\cite{real2017large} approaches.
Additionally, this problem can be more significant for object detection than image classification, due to the more complicated pipelines with larger input images.

To this end, we propose a computation-friendly method, named Fast And Diverse (\name), to search for the task-specific sub-networks in one-stage object detectors.
\name consists of a designed search space and an efficient search algorithm. 
We first design a rich set of diverse transformations tailored for object detection, covering multiple RFs and various convolution types. To learn the optimal combinations more efficiently, a search method via \emph{representation sharing} (\rep) is proposed accordingly. 
By sharing intermediate representations, the proposed \rep significantly reduces the searching time and memory cost for the architecture search. 
Furthermore, we propose an efficient method to reduce the interference between the transformations sharing the same representations, and at the same time, alleviate the degradation of search quality caused by \rep.

To demonstrate the effectiveness of the proposed method, we redesign the sub-networks for modern one-stage object detectors and propose a searchable module for replacement.
\textit{The architecture search for the module is extremely efficient using our \name, 
which is more than $25\times$ faster than the fastest NAS approach for object detectors so far, while achieving a comparable AP improvement} (see Table~\ref{tab:gpudays}).
With ResNeXt-101~\cite{xie2017aggregated} as the backbone, our \name detector achieves $46.4$ AP on the MS-COCO~\cite{lin2014microsoft} \emph{test-dev} set using a single model under single-scale testing, without using any additional regularization or modules (\eg deformable conv~\cite{dai2017deformable}).
Moreover, we show that \name can also benefit more challenging tasks, such as instance segmentation.  
The contributions of this work are summarized as:

\begin{itemize} 
\item[$\sbullet$] We present a novel method, named Fast And Diverse (FAD), to search meaningful transformations in the task-specific sub-networks for one-stage object detection.
The search space is designed specifically for object detection, and we empirically investigate the importance of the RFs coverage and convolution types for object detection.

\item[$\sbullet$] We propose an efficient search method with a novel representation sharing (\rep) algorithm, which can significantly reduce the search cost in both time and memory usage, \eg being more than $25\times$ faster than all previous methods.
To ensure the search quality, a new method is introduced to decouple the transformation selection from the shared representations.
\item[$\sbullet$] To evaluate our methods, we design a searchable module for one-stage object detection and instance segmentation. Extensive experiments show that our \name detector obtains consistent performance improvements on different detection frameworks with various backbones, and even has fewer parameters.

\end{itemize}

%% file: tables/gpudays.tex
\begin{table}[t]
\centering
\small
\caption{{\bf Comparison against other NAS methods for object detection on MS-COCO~\cite{lin2014microsoft}.}
\emph{Trans.} indicates the number of transformation types in the search space (`skip-connect' is excluded). 
\emph{Counterpart} denotes the baseline detectors (and backbone) for direct comparison.
$^\star$ means only the dilation rates are varied.
}
\begin{tabular}{l@{~~}@{~~}c@{~~}c@{~~}c@{~~}@{~~}l@{~~}c}
\hline
\multirow{2}{*}{Method}  &  Search  & \multirow{2}{*}{Trans.}  &  \multirow{2}{*}{GPU-days} & \multirow{2}{*}{Counterpart} & Relative \\
 & Method & & & & AP Imp. \\
\hline

NAS-FPN~\cite{ghiasi2019fpn} & RL & 2 & $>100$ & RetinaNet (Res-50)  & $\uparrow 2.9$  \\ 
DetNAS~\cite{chen2019detnas} &  EA & 4 & 44 & FPN (ShuffleNetv2) & $\uparrow 2.0$ \\ 
NATS-det~\cite{peng2019efficient}  & EA &  $\;\,9^\star$  & 20  & RetinaNet (Res-50) & $\uparrow 1.3$ \\ 
Auto-FPN~\cite{xu2019auto}   & Gradient & 6 & 16 & FPN (Res-50) & $\uparrow 1.9$ \\ 
NAS-FCOS~\cite{wang2019fcos} & RL & 6 & 28 & FCOS (Res-50) & $\uparrow 1.7$ \\ 
SM-NAS~\cite{yao2019sm} & EA & - &  $>100$ & - & -\\
\hline
\name (ours)  & Gradient & 12 & \textbf{0.6} & FCOS (Res-50) & $\uparrow 1.7$ \\

\hline

\end{tabular}

\label{tab:gpudays}
\end{table}

%% file: related.tex
\section{Related Work}

\subsection{Object Detection and Instance Segmentation}
In general, object detectors can be categorized into two groups: two-stage detectors and one-stage detectors. Modern two-stage detectors~\cite{ren2015faster,dai2016r}
first adopt a regional proposal network (RPN) to generate a set of object proposals, which are then fed to the R-CNN heads for object classification and bounding box regression.
On the other hand, one-stage object detectors~\cite{redmon2018yolov3,liu2016ssd,lin2017focal}
directly perform object classification and box regression simultaneously at each spatial location on the feature maps produced from a backbone network. Taking RetinaNet as an example, it consists of a backbone network with a feature pyramid network (FPN) ~\cite{lin2017feature} and two sub-networks for classification and bounding box regression. 
Recent works attempt to get rid of hand-designed anchor boxes while achieving comparable performance~\cite{law2018cornernet,duan2019centernet,zhu2019feature,tian2019fcos}. For instance, FCOS~\cite{tian2019fcos} additionally predicts a centerness score which indicates the distance of current location to the center of the corresponding object,
and can even outperform RetinaNet.

\paragraph{Receptive fields (RF).}
RF is proved to be very important for object detectors~\cite{liu2018receptive,li2019scale}. For instance, Liu \etal \cite{liu2018receptive} designed a combination of kernel sizes and dilation rates, to simulate the impact of the eccentricities of population receptive fields in human visual cortex. 
TridentNet~\cite{li2019scale} tackles the scale variation using multi-branch modules with different dilation rates.
In this work, we aim to search for an optimal combination of different conv layers and dilation rates jointly.

\paragraph{Instance segmentation.}
Instance segmentation is closely related to object detection, and the dominant instance segmentation methods often have two stages~\cite{he2017mask}: they first detect the objects in an image, and then predict an object mask on each detected region. 
Mask R-CNN~\cite{he2017mask} is a representative work in this paradigm, which has an additional mask head on top of Faster R-CNN~\cite{ren2015faster} to perform mask prediction on each object proposal.
In this work, we apply the proposed FAD search method to instance segmentation, which has not been explored previously.

\subsection{Neural Architecture Search}
Recent attention has been moved from network design by hand to neural architecture search (NAS)~\cite{zoph2016neural,pham2018efficient,liu2018progressive,luo2018neural,liu2018darts}. 
A stream of efficient NAS methods is the differentiable NAS~\cite{luo2018neural,liu2018darts}. In particular, DARTS~\cite{liu2018darts} significantly increases the search efficiency by relaxing the categorical choice of operation to be continuous, so that the architecture can be optimized by gradient descent.
In this work, we develop an efficient NAS algorithm for object detectors, by fast searching the optimized transformations.

\paragraph{NAS for Object Detection}
NAS has been applied to many vision tasks apart from image classification, such as object detection~\cite{ghiasi2019fpn,chen2019detnas,xu2019auto,peng2019efficient}. For example, NAS-FPN~\cite{ghiasi2019fpn} uses a RL-based NAS to search for an optimal FPN~\cite{lin2017feature} on the RetinaNet.
DetNAS~\cite{chen2019detnas} aims at finding the optimal shuffle-block-based backbone network in object detectors using an evolution algorithm~\cite{goldberg1991comparative,real2019regularized}. 
A channel-level NAS is proposed in NATS~\cite{peng2019efficient} to search for the backbone in object detectors.
Alternatively, some recent works search for the detection-specific parts rather than the backbone for object detection.
For instance, Auto-FPN~\cite{xu2019auto} searches for a FPN structure
and head structures. 
SM-NAS~\cite{yao2019sm} also searches for two-stage detectors by first conducting a  structural-level search and then a modular-level search. Instead of exploring novel structures, CR-NAS~\cite{liang2019computation} aims to re-allocate the computation resources in the backbone.
NAS-FCOS~\cite{wang2019fcos} is a FCOS-based detector in which the structure of its FPN and the following sub-networks are computed using RL-based NAS.
In this work, we design the search space specifically for object detection, and propose the \name method to search for the sub-networks in one-stage detectors.

%% file: searchSpace.tex
\section{Fast Diverse-Transformation Search}

\subsection{Search Space of \name}
\label{sec:space}

\begin{figure*}[t!]
\centering
\includegraphics[width=0.95\linewidth]{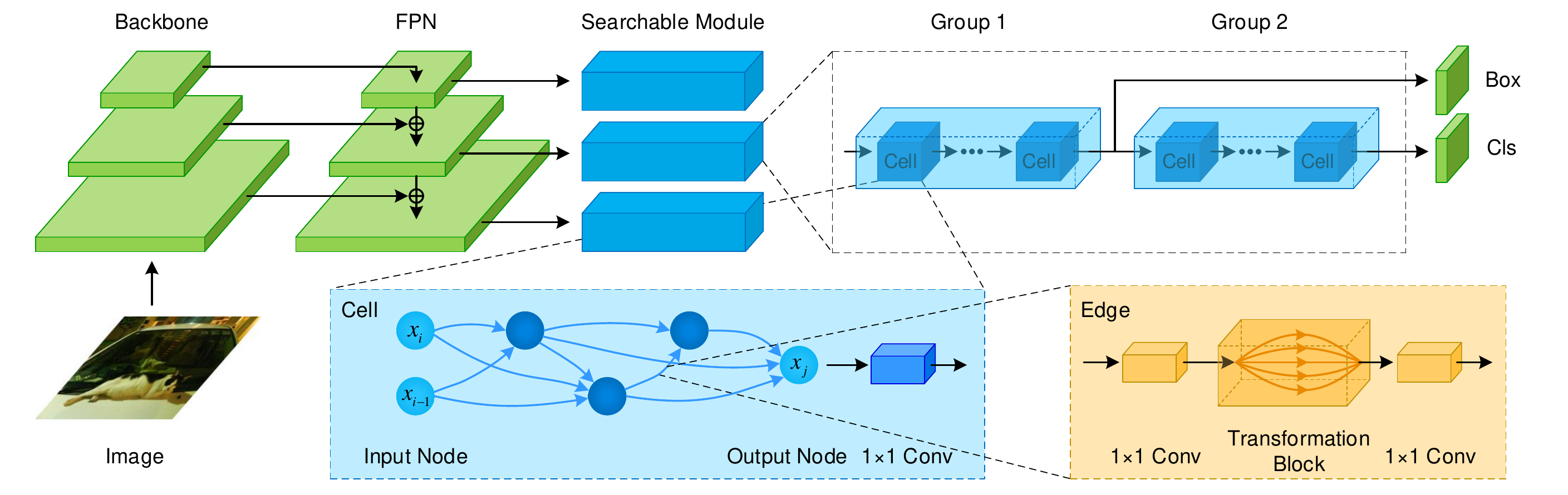}
\caption{ {\bf Search space of \name for one-stage object detectors.} 
The backbone and FPN~\cite{lin2017feature} in detectors 
remain the same, while each FPN level is connected to a searchable module. It consists of two groups of cells, with same cell architectures within each group. In a cell, the edges connecting nodes consist of two standard $1\times1$ conv layers and a transformation block in between. 
The cell structures and the transformations  are to be searched. 
Each edge might have different RFs, resulting in combinations of RFs at each node which enrich the features for capturing information of various scales.
}
\label{fig:detector}
\end{figure*}

One-stage detectors like RetinaNet~\cite{lin2017focal} and FCOS \cite{tian2019fcos} consist of a backbone network with FPN~\cite{lin2017feature} and two parallel sub-networks  
for object classification and bounding box regression, respectively.
In this section, we design a searchable module to replace the commonly-used sub-networks. This module is searched by the proposed \name, and  can be adapted to one-stage object detectors that follow a similar structure as RetinaNet~\cite{lin2017focal} in a plug-and-play fashion.
We then describe the novel search space of \name which is tailored for object detection, including a variety of diverse transformations with different RFs.

\subsubsection{Object Detector with \name}

As shown in Figure~\ref{fig:detector}, the proposed searchable module is comprised of two groups of cells, which are connected sequentially with a shortcut from the input of the module to that of the second group. The module outputs both  object classification and bounding box prediction. The architectures and parameters are shared across different FPN levels.

\paragraph{Classification and regression.}
In \name, the bounding box prediction is performed on the output of the first group, while the classification is computed from the output of the second group. The intuition behind this design is that the two tasks should not be implemented on exactly the same feature maps due to different objectives:  bounding box regression needs to focus on the local detailed information, while object classification is implemented on the features with more semantic information (\ie the feature maps on deeper layers). Therefore, we perform bounding box regression on the output of the first group.

\subsubsection{Design of Search Space}
\label{subsec:ss}
In the following, we describe the design of the search space for \name, which is inspired by the insights from  modern neural architectures~\cite{szegedy2015going,he2016deep} and object detectors~\cite{liu2018receptive,li2019scale}. Three important considerations in our design are the  coverage of RFs, the diversity in convolution types and the computational efficiency.

\paragraph{Groups and cells.}
A group contains $M$ repeated cells, and each cell is defined as a module that contains multiple nodes and edges. Similar to~\cite{liu2018darts,liu2019auto}, each cell is formulated as a directed acyclic graph of nodes. Each node is a stack of feature maps and each edge is an atomic block for search. In this work, we empirically set the number of nodes in each cell to be 3, excluding the input and output nodes.
In our design, an edge consists of two $1\times1$ conv layers $f$ and a transformation block $T$ between the two (Figure~\ref{fig:detector} bottom-right). 
The transformation block contains a set of candidate transformations which will be described in Sec.~\ref{sec:sharing}. Each conv layer in the transformation is followed by a group-normalization layer~\cite{wu2018group} and a ReLU.
Given a node $x_j$, all the predecessors $x_i$ connected to it, and an edge pointing from $x_i$ to $x_j$, we can have the following expression:
\begin{equation}
    x_j = \sum^N_{i<N} f^{c',c}_{i,j}(T^{c',c'}_{i,j}(f^{c,c'}_{i,j}(x_i))),
    \label{eq:node}
\end{equation}
where $f^{c,c'}_{i,j}$ and $f^{c',c}_{i,j}$ are the two $1\times1$ conv layers, with one transforming the input channel $c$ to the channel used in the transformation block $T^{c',c'}_{i,j}$ and the other vice versa. $x_j$ is computed based on $N$ total number of predecessors.
The two $1\times1$ convolution enable flexibility in the channel size in $T$, similar to the inception module~\cite{szegedy2015going}, while maintaining the same channel size for all the nodes. We empirically found that maintaining a relatively large channel size for nodes is beneficial to the performance.
The representations of the intermediate nodes in a cell are concatenated and passed to a $1\times1$ conv layer to reduce the number of channels back to $c$. This additional conv layer ensures the consistent channel size between the input and output of each cell.
Furthermore, the idea of having two groups of cells enables larger flexibility for the architecture search, \ie a larger search space.
Within each group, the cells share the same structure. Therefore, once the search is completed, the cells in each group can be repeated for multiple times, offering great scalability in architecture depth.

\paragraph{Diverse transformations.}
Our initial design of the candidate transformations covers 4 different sizes of RFs (Figure~\ref{fig:space} bottom left).
In particular, for the transformations that are responsible for a RF larger than 5, we use more efficient operations by having a base filter followed by a dilated convolution which spreads out the base filter to reach larger RFs. Moreover, the dilated conv layers are depthwise separable~\cite{sifre2014rigid,howard2017mobilenets}, in order to keep the computation efficient.
The memory-efficient design introduced in Section~\ref{sec:sharing}, allows us to include more types of convolutions. Hence we have two streams of transformations: the standard conv and the depthwise separable conv. 
Namely, for the 6 transformations shown in the bottom-left corner of Figure~\ref{fig:space}, the `conv' layers can be all standard convolution or depthwise separable convolution.

There are no pooling layers involved in the search space as we empirically found that they are not helpful in our scenario. This is probably because the spatial resolution of the feature maps remains the same in the sub-networks.
Moreover, skip-connection is not included in the transformation.
Lastly, a `none' path, indicating the importance of input edges with respect to each node, is added to the transformation block.
In summary, the proposed transformation block contains 13 distinct transformations in total, including 2 types of conv layers and 3 dilation rates, and covering 4 sizes of RFs, as illustrated in Figure~\ref{fig:space}.
Therefore, we build a meaningful search space with strongly diverse transformations.
The resultant search space has roughly $2.3\times10^{13}$ unique paths in total, with one cell per group in search time.

\subsubsection{\name for Instance Segmentation}
\label{sec:space:mask}

We expect that the mask prediction task can also benefit from the combination of RFs and diverse transformations. 
With minimal modification, \name readily applicable to general instance segmentation frameworks, \eg Mask R-CNN~\cite{he2017mask} and Mask Scoring R-CNN~\cite{huang2019mask}. 
Specifically, we replace the conv layers before the deconvolutional layer in the mask head by the proposed searchable module, and search for its architecture in an end-to-end fashion. The search space follows that of object detection.

%% file: sharing.tex
\subsection{Fast Search with Representation Sharing}
\label{sec:sharing}

In this section, we propose a novel algorithm to significantly reduce the search cost in both time and memory, followed by the description of the search procedure.

\subsubsection{Representation Sharing}
\label{subsec:sharing}

The proposed acceleration method for architecture search, named \rep, is performed in two steps: filter decomposing and intermediate representation sharing. We elaborate on these two steps in the following.

\begin{figure*}[t]
\centering
\includegraphics[width=0.999\linewidth]{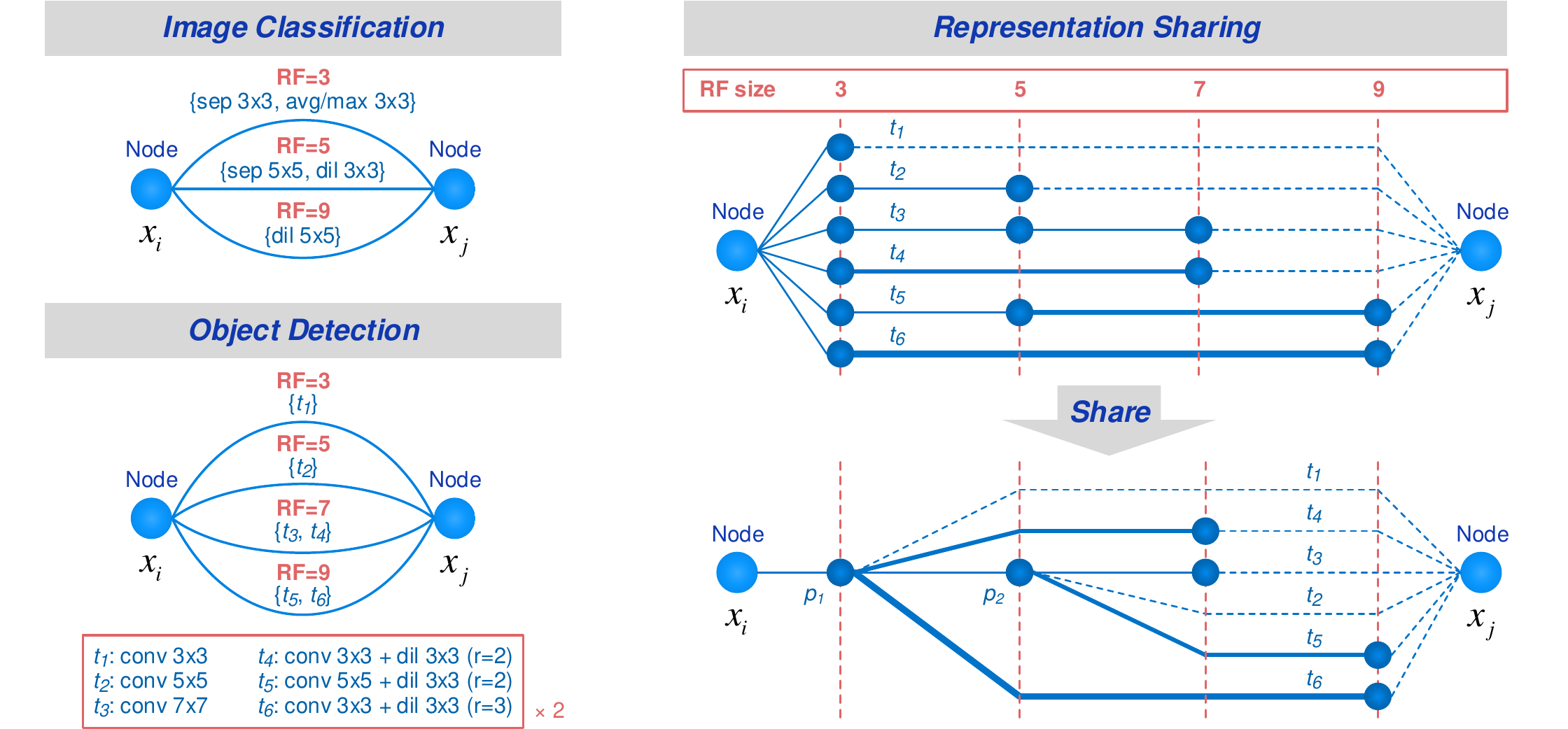}
\caption{ {\bf Transformations and representation sharing.}
\textbf{Left}: comparison between the transformations used for image classification and those proposed for object detection in the search space. The proposed transformations are listed at the bottom. \emph{conv} can be the standard or the depthwise separable convolution.
\textbf{Right}: \rep. 
Each sphere and solid line denotes a representation and a conv layer, respectively.
First, large filters are decomposed into stacks of $3\times3$ filters. Second, \emph{$p_1$} and \emph{$p_2$} are shared across transformations.
Note that the $1\times1$ conv layers are not shown for simplicity.
}
\label{fig:space}
\end{figure*}

\paragraph{Decomposing large filters.}
As proposed in~\cite{simonyan2014very}, filters with large kernel sizes can be replaced by multiple $3\times3$ filters. For example, a stack of three $3\times3$ filters in fact has an equivalent size of receptive filed as a $7\times7$ filter.
The stacked filters have the advantages of fewer parameters and more non-linearities in between for learning more discriminative representations.
Following this intuition, we decompose the filters with large kernel size and construct a transformation block only containing filters of size $3\times3$ (\emph{$t_1$} to \emph{$t_6$} shown in Figure~\ref{fig:space} top-right).
However, the replacement of large filters with stacks of small ones significantly increases the memory overhead during the search.
Taking the proposed transformations as an example, more than twice intermediate representations are generated after the decomposition.

\paragraph{Representation Sharing.}
To reduce this memory overhead, we further propose a novel approach.
Namely, for each receptive field (RF) level, all the intermediate representations that are not directly connected to node $x_j$ are shared (Figure~\ref{fig:space} bottom-right).
To be specific, we denote \emph{$t_3$} in top-right of Figure~\ref{fig:space} as the stem. In the stem, there are 3 intermediate representations having different sizes of RFs with respect to the node $x_i$. 
We merge the transformations by sharing the intermediate representations in the stem. 
For example in Figure~\ref{fig:space} (top-right), to merge the $t_1$ into the stem, we directly connect the first intermediate representation in the stem to node $x_j$, and therefore the original $t_1$ (conv $3\times3$) transformation is replaced by this new transformation. 
Specifically, the \rep reduces the number of representations computed in each transformation block from 26 to 12. 
Therefore, it can significantly speed up the search process.
Moreover, the search speed is further boosted by the memory-efficiency of \rep since the search can be done using a single GPU, which avoids the computational overhead introduced by training with multiple GPUs (e.g. parameter update).

\paragraph{Relation to other efficient search methods.}
The proposed \rep has similar spirits to some recent approaches.
For instance, parameter sharing introduced in~\cite{pham2018efficient} takes advantage of sharing the same sets of parameters among child models to greatly speed up the search in RL-based NAS methods. It is inspired by parameter inheritance~\cite{real2017large} which also reuses the same parameters for child models across mutation to avoid training from scratch. 
\rep is more than using the same parameters, but also the same computation. Furthermore, apart from accelerating the search, \rep further reduces the memory consumption. 
Single-path NAS~\cite{stamoulis2019single} also share computations, but is different from ours. It considers a small kernel (\eg $3\times3$) as the core of a large one (\eg $5\times5$), and uses a learnable threshold to compare the importance of the two kernels, and selects the optimal one.

\subsubsection{Decoupling Shared Representations}
Similar to parameter sharing described in~\cite{pham2018efficient} in which child models are coupled to some extent due to reusing the same weights, \rep also exhibits such behaviour. In \rep, transformations sharing the same representations might interfere with each other, and thus the parameters directly corresponding to the shared representations are not well optimized in the search. 
It causes that those transformations are difficult to outstand in the architecture derivation.
For example, in Figure~\ref{fig:space} (bottom-right), two intermediate representations are shared. Namely, $p_1$ is shared across all six transformations and $p_2$ is shared across $t_2$, $t_3$ and $t_5$. Due to the coupling effect (\ie interference between transformations), $t_1$ and $t_2$ are not able to learn the optimal parameters on their own, which may degrade the search quality.
Notably, this effect mainly happens on $t_1$ and $t_2$, since their outputs are exactly the shared representations; while other transformations ($t_3$ to $t_6$) have the flexibility to compensate this effect due to additional operations on the share representations.

\paragraph{Decoupling with extra functions.}
To address this issue in \rep, we propose a simple yet effective method to decouple the transformations (that directly depend on the shared representations, \ie $t_1$ and $t_2$) from the shared representations. Namely, an additional function $H$ is applied between each shared representation and its corresponding transformation output. With this additional function, for example, the output of $t_1$ is no longer $p_1$, but $H(p_1)$. In this case, $t_1$ and $t_2$ are decoupled from $p_1$ and $p_2$, respectively.  
For the choice of $H$, we use a standard $1\times1$ conv layer followed by a ReLU. This light-weight extra function produces minimal computational overhead and is applied to both conv streams (\ie the standard and depthwise separable convolution streams).

\subsubsection{Optimization and Deriving Architectures}
\label{sec:optimization}

In a cell, each edge contains a transformation block in which the final transformation is determined from a set of candidates illustrated in Figure~\ref{fig:space}. 
In order to search using back-propagation, we follow the continuous relaxation for the search space as~\cite{liu2018darts}, and adapt it to the proposed \rep paradigm.
For each of the two streams (Figure~\ref{fig:space} bottom-right) in the transformation block, the output of a transformation ($T_{i,j}$) is essentially the sum of all the intermediate representations multiplied with corresponding $\alpha$. Therefore, we can have:
\begin{equation}
    T_{i,j}(x_{i}^{\prime})=\sum_{p \in P} \frac{\exp \left(\alpha^p_{i,j}\right)}{\sum_{p^{\prime} \in P} \exp \left(\alpha^{p^{\prime}}_{i, j}\right)} \; p,
    \label{eq:relax_new}
\end{equation}
where $x_i^{\prime}$ is the output of the first $1\times1$ conv layer in the transformation block. 
$p$ and $p^{\prime}$ are the intermediate representations out of all representations $P$.
$\alpha^p$ is the $\alpha$ corresponding to $p$.

\paragraph{Optimization and derivation of discrete architectures.}
During the architecture search, $\alpha$ and the network weights $w$ are jointly optimized in a bilevel optimization scheme, as in~\cite{liu2018darts,liu2019auto}. 
In particular, the first-order approximation is adopted.
At the end of the search, 
a discrete architecture is decoded
by retaining one transformation per edge and two input edges for each node based on the largest $\alpha$ in each transformation block. 
Since the intermediate representations are selected instead of operations, they should then be mapped to the corresponding actual transformations in the derived architecture, \ie the transformations in Figure~\ref{fig:space} (top-right).

%% file: exp.tex
\section{Experiments}

In this section, the proposed \name is evaluated in two tasks: object detection and instance segmentation. In 
Section~\ref{sup:classification}, we further conduct experiments for image classification to analyze the effect of decoupling in \rep.

\subsection{Object Detection}
\label{sec:exp:detect}

\paragraph{Implementation details.}
Although the proposed module can be adopted in different one-stage object detectors, we perform the architecture search using \name on FCOS~\cite{tian2019fcos}, due to its efficiency.
The search is conducted on the PASCAL VOC~\cite{everingham2010pascal}.
We also perform the search directly on MS-COCO~\cite{lin2014microsoft} and make comparisons in Table~\ref{tab:space}.  
More implementation details, including the search and the detector training, can be found in the Section~\ref{sup:detect}.

\subsubsection{Ablation Study}
\label{subsec:ablation}
We conduct ablation study on the search cost, search spaces, as well as different backbones and detectors. 
More studies on the marco-structure of the module, and network width and depth are presented in Section~\ref{sup:detect}.

\paragraph{Search cost.} 
The time required for a complete architecture search using our \name is $0.6$ GPU-days. A single TITAN XP is used for the search. Table~\ref{tab:gpudays} compares the search cost of \name against other NAS-based methods for object detection. As we can see, the search speed for \name is at least $25\times$ faster than other recent approaches, while achieving a similar relative AP improvement on MS-COCO. Meanwhile, the  architecture explored by \name is scalable in depth by simply adding the repetitive cells in the groups, which provides greater flexibility to the module.

\input{tables/space.tex}

\paragraph{Search space.}
To demonstrate the superiority of the proposed search space, we reuse the same search procedure but replace the proposed search operations with that in DARTS~\cite{liu2018darts}, which are listed in Figure~\ref{fig:space} (top-left). Note that the depthwise separable convolution is doubled in DARTS, and hence the RFs change accordingly.
As we can see from Table~\ref{tab:space}, the operations used in DARTS only bring a marginal improvement of $0.4$ AP, compared to the original FCOS, while \emph{the proposed transformations improve the performance significantly, from $38.6$ to $40.3$}. 
To further study the importance of the full transformation set, we search by using two transformation subsets. Namely, the two subsets contain transformations with the RFs smaller than $7$ and $9$, respectively. Our results show that with fewer transformations in the search space, the performance  degrades accordingly.
Moreover, we search by using only one type of convolution (either the standard or the depthwise separable) for the conv layers with dilation rate of 1. 
Not surprisingly, both of them fail to achieve a similar performance as the full search space.
This illustrates the power of the proposed transformations which fully benefit from the better combinations of RFs and convolution types.
Besides, the performance slightly degrades without decoupling. 
More results on decoupling can be found in the SM.
Another observation is that the proxyless search on MS-COCO can achieve similar performance on detection, but it takes much longer search time. Hence, we use the architecture searched on VOC for object detection for the rest of this work.

Additionally, our \name is also compared with the `random' baseline. Namely, a transformation is randomly sampled in each block and two edges are randomly sampled for each node. It can be found that the proposed \name indeed finds much better architectures. The last conclusion to draw in Table~\ref{tab:space} is that comparing to the search without \rep (using 4 GPUs), \emph{\rep enables an almost  $4\times$ faster search with only one GPU}, without harming the performance.

\paragraph{Adaptation to different backbone networks.}
We replace the ResNet-50 in the detector by using three different networks: MobileNetV2~\cite{howard2017mobilenets}, ResNet-101~\cite{he2016deep} and ResNeXt-101~\cite{xie2017aggregated}. As shown in Table~\ref{tab:params}, our \name obtains a consistent improvement (about $1.4$ AP on average) for all the backbones compared, with even fewer parameters and FLOPs. This indicates that the architecture of \name generalizes well to the backbone networks with different capacity.
Direct comparison on the sub-networks (without the backbone and FPN) shows a $16.3\%$ and $15.2\%$ decrease on the number of parameters and the FLOPs.
Hence, we can conclude that the performance gain is obtained from the better architecture searched rather than the network capacity itself.

\input{tables/param.tex}

\paragraph{Transferability.}
Our \name is expected to be readily applicable to different types of one-stage object detectors (with the two-subnet structure). To examine this property, we further plug the proposed searchable module into RetinaNet~\cite{lin2017focal}.  Table~\ref{tab:params} reveals that \name can also improve the performance of RetinaNet by a large margin even with fewer parameters. 
Therefore, we see that the searched  sub-networks can boost the performance of different types of  detectors (and potentially more powerful detectors in the future) in a plug-and-play fashion.

\subsubsection{Comparison with the state-of-the-art}
\label{sec:sota}

\input{tables/sota.tex}

We compare \name with the state-of-the-art object detectors on the MS-COCO \emph{test--dev} split, including some recent NAS-based object detectors. All the methods are evaluated under the single-model and single-scale setting.
Table~\ref{tab:sota} shows that, by having 128 channels in the first group and 256 in the second (with $98.3$M parameters), \name @128-256 achieves $46.4$ AP which surpasses all the recent object detectors, including two concurrent work, NAS-FCOS~\cite{wang2019fcos} and Hit-Detector~\cite{guo2020hit}.
Note that 
NAS-FCOS includes the deformable convolution~\cite{dai2017deformable} in the search space, which is not considered in other NAS-based detectors (including our \name), and it is well-known for giving large AP improvements. 
On the other hand, the search of \name is almost $50\times$ faster than that of NAS-FCOS on the same dataset (\ie VOC).

\subsubsection{Searched Architectures}

\begin{figure*}[t]
\centering
\includegraphics[width=0.95\linewidth]{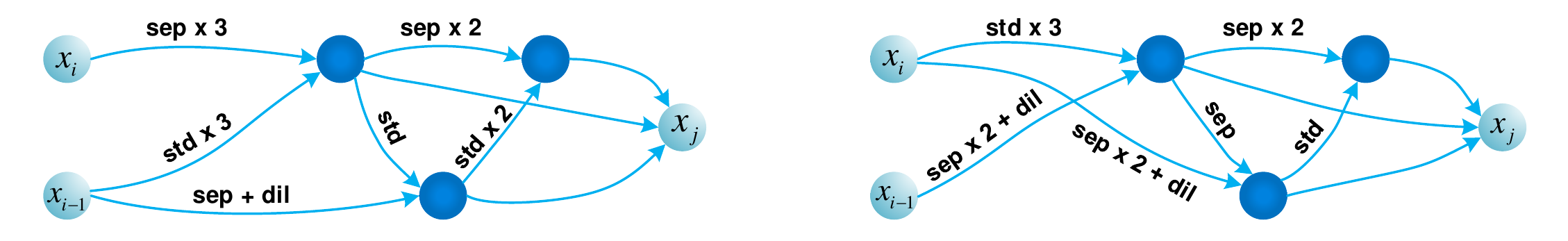}
\caption{ {\bf Architectures searched for object detection. }
The left and right cells are for the first and second group, respectively. \emph{std}, \emph{sep} and \emph{dil} denote the standard, depthwise separable and dilated conv.
}
\label{fig:geno}
\end{figure*}

The derived architectures by \name  are presented in Figure~\ref{fig:geno}.
We have two interesting observations. First, the edges correspond to a mixture of RFs (especially for the cell group for classification) and convolution types, which again validates our motivation.
Another important insight is that the transformations with large RFs (\ie 7 and 9) appear near the input node, while those with small RFs (\ie 3 and 5) are closer to the output node. This architecture is consistent with the DetNAS architecture explored in~\cite{chen2019detnas}.
Notably, architectures better than the ones shown in Figure~\ref{fig:geno} may be found in the search space if the search is repeated for more times.

\subsection{Instance Segmentation}
\label{sec:mask}

To showcase the generality of the proposed \name, we apply it to another useful task -- instance segmentation.
Different from object detection, only one group of cell is searched in the mask head. 
The search is conducted on MS-COCO, which takes 2.6 GPU-days.
For a fair comparison, we exactly follow~\cite{he2017mask,huang2019mask} for training the searched networks. The search and training details are described in the Section~\ref{sup:mask}.

\subsubsection{Results}

\input{tables/mask}

Table~\ref{tab:mask} shows that, with a similar number of parameters and FLOPs, all \name outperform their counterparts with the same backbones on both Mask R-CNN and MS R-CNN. Notably, Mask \name has relatively larger improvements in terms of $\mathrm{AP_{M}}$ and $\mathrm{AP_{L}}$ (\eg 1.6 and 1.8 AP on ResNet-50) than $\mathrm{AP_{S}}$ (0.6 AP), possibly due to the transformations with larger RFs.
Another surprising result is that Mask \name (ResNet-50) achieves similar AP as MS R-CNN (ResNet-50), \ie $35.5$ \vs $35.6$, despite a simpler pipeline and $26.9\%$ fewer parameters.
The improvements are prominent since we only modify the mask head architecture which only accounts for 2.25M parameters, \ie 2.8\% to 5\% of the whole networks.
The explored cell architecture for the mask head is displayed in Section~\ref{sup:mask}.

%% file: tables/space.tex
\begin{table}[t]

\caption{{\bf Comparison for the architecture search.}
\emph{Memory} and \emph{bs} denotes the memory usage and images per GPU.
Both \emph{Subset} and \emph{Full} refer to the proposed search space.
\emph{Sep.} and \emph{Std.} mean that only depthwise and standard conv are used, respectively.
ResNet-50 is used as the backbone. Results are obtained on the MS-COCO \emph{minival} split.
All the searches are performed on VOC, expect for $^{\dagger}$ which is on MS-COCO.}

\centering

\begin{tabular}{lc@{~}c@{~~}c@{~~}c@{~~}cc@{~~}c}
\hline
Method  &   \rep  &  Search Space    & Trans.  & RFs   & Memory (G)  & GPU-days    & AP  \\ \hline

FCOS~\cite{tian2019fcos} & - & -  & - & - & -   & - & 38.6    \\ 
Random   & -  & Full  & 12 & {3,5,7,9} & -  &  -  & 39.0  \\
\hline

\name   &  \xmark  & DARTS~\cite{liu2018darts}  &  7 &  {5,7,9} & $\sim10$ ($bs=4$) & 0.4 & 39.0   \\

\name    & \cmark & Subset 1  & 4 & {3,5} & $\sim7$ ($bs=4$)  & 0.25 & 39.2  \\
\name    & \cmark & Subset 2  & 8 & {3,5,7} & $\sim11$ ($bs=4$)  & 0.5  & 39.7 \\

\name    & \cmark & Sep. only  & 6 & {3,5,7,9} & $\sim10$ ($bs=4$)  & 0.36 & 39.5 \\ 
\name    & \cmark & Std. only  & 6 & {3,5,7,9} & $\sim9.5$ ($bs=4$)  & 0.4 & 39.9 \\ 

\name    & \cmark & w/o decouple & 12 & {3,5,7,9} & $\sim12$ ($bs=4$)  & 0.6 & 40.0 \\ 
\name    & \cmark & Full  & 12 & {3,5,7,9} & $\sim12$ ($bs=4$)  & 0.6 & \textbf{40.3} \\ 
\name   & \xmark & Full  & 12 & {3,5,7,9} & $\sim9$ ($bs=1$)  & 2.3 & \textbf{40.3} \\
\hline

\name$^\dagger$   & \cmark & Full  & 12 & {3,5,7,9} & $\sim9$ ($bs=4$)  & 5.5 & \textbf{40.3} \\

\hline

\end{tabular}

\label{tab:space}
\end{table}

%% file: tables/param.tex
\begin{table}[t]
\centering
\small
\caption{{\bf \name on different detectors and backbones.}
The $\rightarrow$ indicates the change from original detector to \name.
\emph{Dim.} is the channel size in the subnets, or $c^{\prime}$ in the transformation block in \name.
Results are obtained on MS-COCO \emph{minival}.
}

\begin{tabular}{l@{~~}@{~~}l@{~~}l@{~~}@{~~}c@{~~~~}c@{~~~~}c}
\hline
Method  & Backbone  & Dim.   &    Params (M)  & FLOPs (G)     & AP   \\ \hline 

\multirow{5}{*}{FCOS} & MobileNetV2  & $256 \rightarrow 96$ &   $9.8 \rightarrow 9.0$ & $124 \rightarrow 108$ & $31.3 \rightarrow 32.7$    \\
 &  Res-50  &  $256 \rightarrow 96$ & $32.2 \rightarrow 31.5$ & $201 \rightarrow 185$ & $38.6  \rightarrow 40.3$ \\
 &  Res-101  & $256 \rightarrow 96$  &   $51.2 \rightarrow 50.4$ & $277 \rightarrow 261$ & $43.0 \rightarrow 44.2$   \\
 & Res-X-101 & $256 \rightarrow 96$ & $90.0 \rightarrow  89.2$ & $439 \rightarrow 423$ & $44.7 \rightarrow 45.8$ \\
 & Res-X-101  & $256 \rightarrow 128$  &    $90.0 \rightarrow  91.2$ & $439 \rightarrow 465$ & $44.7 \rightarrow \textbf{46.0}$   \\
\hline

\multirow{3}{*}{RetinaNet}  & Res-50 & $256 \rightarrow 96$ &  $33.8 \rightarrow 33.0$ & $234 \rightarrow 218$ & $36.1 \rightarrow  37.7$ \\
 &  Res-101 & $256 \rightarrow 96$    &  $52.7 \rightarrow  52.0$ & $310 \rightarrow 294$ & $37.7 \rightarrow 39.4$ \\
 &  Res-X-101 & $256 \rightarrow 128$ &   $91.5 \rightarrow  92.7$ & $472 \rightarrow 498$ & $39.8 \rightarrow \textbf{41.6}$ \\ 

\hline

Subnet only & - & $256 \rightarrow 96$ & $4.9 \rightarrow 4.1$   & $105 \rightarrow 89$  & -  \\

\hline

\end{tabular}
\label{tab:params}
\end{table}

%% file: tables/sota.tex
\begin{table*}[t]
\centering
\small

\caption{{\bf Comparison with the state-of-the-art object detectors} on the MS-COCO \emph{test-dev} split (including concurrent work~\cite{guo2020hit,zhu2019soft,zhang2020bridging,wang2019fcos}).
FCOS~\cite{tian2019fcos} is used as the base detector for \name.
All the results are tested under the single-scale and single-model setting. Note that models using additional regularization method~\cite{ghiasi2018dropblock} and deformable convolution~\cite{dai2017deformable} are excluded in the table (except for NAS-FCOS~\cite{wang2019fcos}). 
}


\begin{tabular}{ll@{~~~}@{~}c@{~~}ccccc}
\hline
\textbf{Two-stage} detectors  & Backbone    & \textbf{AP} & $\mathrm{AP_{50}}$ & $\mathrm{AP_{75}}$ & $\mathrm{AP_{S}}$ & $\mathrm{AP_{M}}$ & $\mathrm{AP_{L}}$ \\ \hline

TridentNet\cite{li2019scale} & ResNet-101 & 42.7  &  63.6  & 46.5 &  23.9  & 46.6  & 56.6 \\
Auto-FPN~\cite{xu2019auto} & ResNeXt-64x4d-101  & 44.3 & -  & - & - & - & -  \\
SM-NAS: E5~\cite{yao2019sm}   & Searched  & 45.9 & 64.6 & 49.6 & 27.1 & 49.0 & 58.0  \\
Hit-Detector~\cite{guo2020hit} & Searched & 44.5 & -  & - & - & - & -  \\
\hline

\textbf{One-stage} detectors & & \\ \hline
RetinaNet~\cite{lin2017focal} & ResNeXt-101  & 40.8 & 61.1 & 44.1  & 24.1 & 44.2 & 51.2 \\
CenterNet511~\cite{duan2019centernet}  &  Hourglass-104  &  44.9  & 62.4  &  48.1  & 25.6 &  47.4 &  57.4  \\ 
FSAF~\cite{zhu2019feature}  & ResNeXt-64x4d-101   & 42.9  & 63.8  & 46.3  & 26.6  & 46.2  & 52.7 \\
FCOS~\cite{tian2019fcos}  & ResNeXt-64x4d-101   & 44.7  & 64.1  & 48.4  & 27.6  & 47.5  & 55.6 \\
FreeAnchor~\cite{zhang2019freeanchor} & ResNeXt-101 & 44.9 & 64.3 & 48.5 & 26.8 & 48.3 & 55.9 \\
SAPD~\cite{zhu2019soft} & ResNeXt-64x4d-101 & 45.4 & 65.6 & 48.9 & 27.3 & 48.7 & 56.8   \\ 
ATSS~\cite{zhang2020bridging} & ResNeXt-64x4d-101 & 45.6 & 64.6 & 49.7 & 28.5 & 48.9 & 55.6   \\ 
NAS-FPN~\cite{ghiasi2019fpn} (7 @ 384) & ResNet-50 & 45.4 & -  & - & - & - & - \\
NAS-FCOS~\cite{wang2019fcos}    & ResNeXt-64x4d-101  & 46.1 & -  & - & - & - & - \\ 
 \hline 

\name @ 96  & ResNet-101   & 44.1 & 62.7  & 47.9 & 26.8 & 47.1 & 54.6 \\
\name @ 128 & ResNet-101   & 44.5 &  63.0 & 48.3  & 27.1 & 47.4 & 55.0  \\
\name @ 128 & ResNeXt-64x4d-101   & 46.0 & 64.9 & 50.0  & 29.1  & 48.8 & 56.6  \\
\name @ 128-256 & ResNeXt-64x4d-101   & \textbf{46.4} & 65.4 & 50.4  & 29.4  & 49.3 & 57.4  \\

\hline

\end{tabular}

\label{tab:sota}
\end{table*}

%% file: tables/mask.tex
\begin{table}[t]
\centering
\small
\caption{\textbf{Comparison on instance segmentation} \emph{mask} AP on the MS-COCO \emph{minival} split.
\emph{P.} is for parameters (M) and \emph{F.} is for FLOPs (G).
}


\begin{tabular}{l@{~~}l@{~~}c@{~~}@{~~}c@{~~}@{~~~}c@{~~~}c@{~~}c@{~~}@{~~}c@{~~}c@{~~}c@{~~}}
\hline
Method  & Backbone     &    P.  & F.     & \textbf{AP} & $\mathrm{AP_{50}}$ & $\mathrm{AP_{75}}$ & $\mathrm{AP_{S}}$ & $\mathrm{AP_{M}}$ & $\mathrm{AP_{L}}$  \\ \hline

\multirow{2}{*}{Mask R-CNN~\cite{he2017mask}} &  Res-50  &   $44.3$ & 285 & $34.2$ & 55.7 & 36.3 & 15.4 & 36.8 & 50.9 \\
 &  Res-101  &  63.3 & 362 &  36.1 & 58.1 & 38.3 & 16.4 & 38.9 & 53.4 \\
\multirow{2}{*}{Mask \name} & Res-50  &  44.4 & 287 & 35.5 & 56.8 & 37.9 & 16.0 & 38.4 & 52.7 \\
 &  Res-101  &  63.4   & 364 & 37.0 & 58.6 & 39.5 & 17.0 & 39.8 & 54.9   \\
\hline 


\multirow{2}{*}{MS R-CNN~\cite{huang2019mask}} &  Res-50  & 60.7  & 326  & 35.6 & 56.2 &  38.2 & 16.6 & 37.8 & 52.0 \\
 &  Res-101  & 79.6    & 402 &  37.4 & 58.3 & 40.2 & 17.5 & 40.2 & 54.4 \\
\multirow{2}{*}{MS \name} & Res-50  & 60.8 & 328 & 36.3 & 56.3 & 39.2  & 16.1 & 38.8 & 53.4 \\
 &  Res-101  & 79.7   & 404  & \textbf{38.0} & 58.7 & 41.0 & 17.6 & 41.0 & 55.1   \\
 

\hline

\end{tabular}
\vspace{-2mm}
\label{tab:mask}
\end{table}

%% file: conclusion.tex
\section{Conclusion}

In this work, we propose  \name to efficiently search for better sub-networks with diverse transformations and optimal combinations of RFs for one-stage object detection and instance segmentation. 
To demonstrate the effectiveness of the proposed search space and search method, we design a searchable module for the two tasks at hand (and potentially applicable to other tasks).
Extensive experiments show that the architectures searched by our \name can consistently outperform their counterparts on different detectors and segmentation networks.

%% file: app.tex
\section{Appendix}

In this appendix, we first
describe the implementation details for object detection, followed by more ablation studies (Section~\ref{sup:detect}). In Section~\ref{sup:mask}, we explain the implementation details for instance segmentation, and then display the explored cell architecture.
Lastly, Section~\ref{sup:classification} provides some details of the search space and search method for image classification, as well as the results on the effect of decoupling \rep.

\subsection{Object Detection}
\label{sup:detect}

\paragraph{Architecture Search for the Subnetworks.}
The search takes $12.5k$ iterations with batch size $4$.
The initial learning rate is $0.004$ and divided by $10$ at iteration 10k. The size of the input images are resized such that the short side is $416$ and the long side is less or equal to $693$. The norm of the gradients are clipped to $20$ to stabilize the search.
During the search, we derive a discrete architecture every 2.5k iterations. The search process is terminated when the current derived architecture is the same as the previous one (\ie 2.5k iterations before it). In our experiments, we find that the search mostly terminates at 12.5k iterations on VOC, and continuing the search does not change the derived architecture. 
Meanwhile, ${L}_{train}(w,\alpha)$, which is also considered as an indicator of the search process, flattens. Therefore, we use 12.5k iterations for all the experiments for consistency. 
To balance the efficiency and consistency between the search and detector retraining, we set $M=1$ and $c^{\prime}=96$ for the search. 
Notably, $c=256$ is used for all the experiments in this work, including the search and training.

\paragraph{Object detector training.}
Once we obtain the derived architectures, we train the whole detector on the MS-COCO \emph{train2017}.
For FCOS, we exactly follow the training strategy as in~\cite{tian2019fcos} for different backbone networks, including input image sizes, learning rate schedule and iterations. Similarly, the same training strategy as in~\cite{lin2017focal} is adopted for RetinaNet.
All the FCOS detectors (including vanilla and \name) are trained using the improvements introduced in~\cite{tian2019fcos}. Note that the centerness in FCOS is predicted based on the first cell group.
For detector training, we set $M=2$ and $c^{\prime}=96$, unless specified.

\subsubsection{More Ablation Studies}

\paragraph{Ablation on the macro-structure of \name.}
Searching and training a \name with two parallel groups for classification and box regression (\ie similar to RetinaNet~\cite{lin2017focal} and FCOS~\cite{tian2019fcos}) achieve $40.0$ AP, which is $0.3$ lower than the sequential-group design in the main paper.
We also experiment with the sequential groups in reverse order, \ie classification is performed after the first group and regression after the second. This results in a drop of $0.1$ AP.
Therefore, the design of the macro-structure brings much smaller improvements comparing to the proposed search method.

\begin{figure}[t]
\centering
\includegraphics[width=0.8\linewidth]{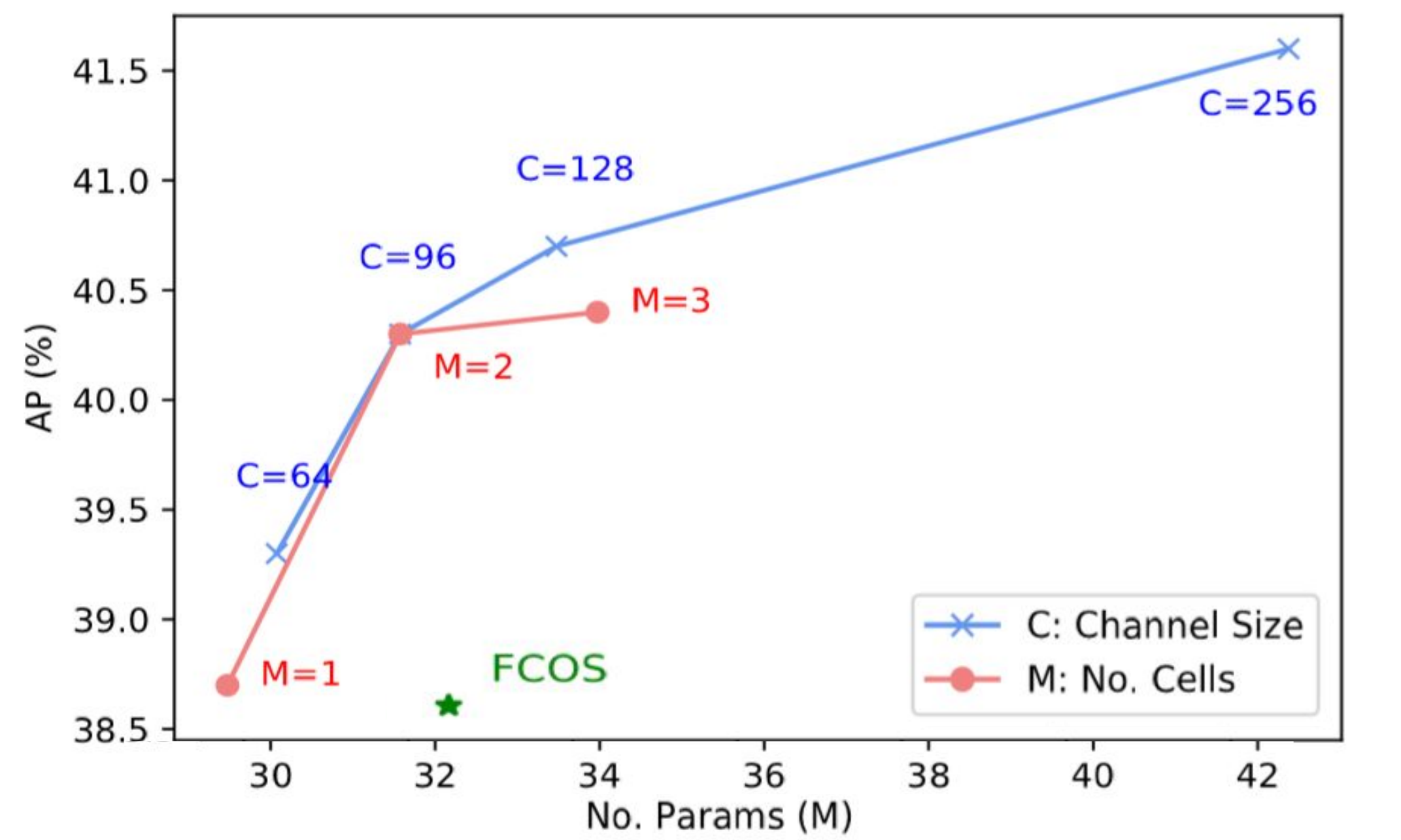}
\vspace{-3mm}
\caption{ {\bf \name with different channel sizes and number of repeated cells.} 
When channel size $C$ varies, the number of repeated cells in the groups $M$ is fixed to be 2; whereas $C$ is fixed to be 96 when $M$ varies.
Results are obtained on the MS-COCO \emph{minival} set, using ResNet-50 as the backbone.
}
\label{fig:ap}
\vspace{-6mm}
\end{figure}

\vspace{-2mm}
\paragraph{Channel size and the number of cells.}
The capacity of the derived architecture searched by \name can vary in two directions: the channel size $c^{\prime}$ in the transformation blocks and the number of repeated cells $M$ in both groups. For a better understanding on their effects, we take \name on FCOS with ResNet-50~\cite{he2016deep} as a base model and vary $c^{\prime}$ and $M$. Figure~\ref{fig:ap} shows the performance trend of both directions. The two curves grow fast at the beginning and then increase in a slower pace, as we expect. Nevertheless, all the \name models outperform FCOS.
As the model with $M=2$ has the best trade-off between performance and model size, we fix $M=2$ throughout the object detection experiments in the main paper.

\vspace{-1mm}
\subsection{Instance Segmentation}
\label{sup:mask}

\paragraph{Architecture search for the mask head.}
The search is performed using Mask R-CNN~\cite{he2017mask} on the MS-COCO \emph{train2017} set, which is randomly split into two halves: one for optimizing the architecture $\alpha$ and the other for learning the network weights $\emph{w}$. 
We set $M$, $c^{\prime}$ and $c$ to be 1, 64 and 256, respectively. 
The search takes 180k iterations, with an initial learning rate of 0.02 and decreased by 10 at the 120k and 160k iteration. The weight decay is $0.00001$.
The rest of search details are the same as object detector search.
The search for mask head architecture requires $2.6$ GPU-days.
We use the same strategy as that for object detectors to derive the searched architecture.

\vspace{-1mm}
\paragraph{Segmentation network training.}
For training Mask \name  with the searched architecture, we exactly follow~\cite{he2017mask}. For a fair comparison, we set $M=2$ and $c^{\prime}=96$, which results in a similar capacity of Mask R-CNN.
We also transfer the searched architecture to a more recent segmentation network, Mask Scoring R-CNN~\cite{huang2019mask} (MS R-CNN). The same training scheme as in~\cite{huang2019mask} is adopted for MS \name (\ie Mask Scoring R-CNN with \name).

\begin{figure*}[t]
\centering
\includegraphics[width=0.60\linewidth]{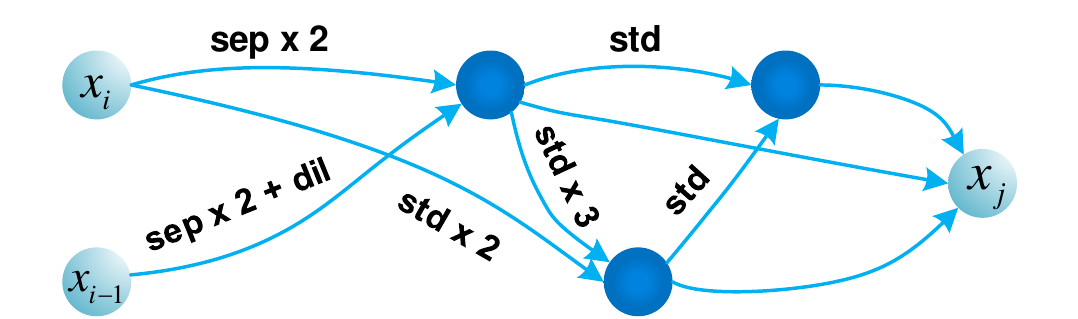}
\caption{ {\bf Architecture searched for instance segmentation. } \emph{std}, \emph{sep} and \emph{dil} denote the standard, depthwise separable and dilated conv, respectively.
}
\label{fig:genomask}
\end{figure*}

\paragraph{Cell architecture for instance segmentation.}
The searched cell architecture using Mask R-CNN is displayed in Figure~\ref{fig:genomask}. This module is used to replace the consecutive conv layers before upsamlping in the mask head.

\subsection{Image Classification}
\label{sup:classification}

To further demonstrate the effectiveness of decoupling the transformations from the shared representations, we conduct ablation study on CIFAR-10~\cite{krizhevsky2009learning}.

\paragraph{Search space and search method.}
We follow the experimental details in~\cite{liu2018darts}, but with a different search space. For normal cells, the candidate operations are $t_1$ to $t_5$, while the reduction cell only  has two options: max/avg-pooling. 
`Skip-connect' is included to both cells. 
Note that the goal of the experiments is not to search for better architectures for image classification. Instead, by constructing this simpler and smaller search space, we can illustrate the decoupling effect more clearly.
To reduce the variance, we search for 4 architectures with or without the decoupling, and train for 5 runs each.

\input{tables/cifar}

\paragraph{Results and discussion.}
Table~\ref{tab:cifar} reports the percentage of two groups of transformations being chosen in the derived architectures and the mean  error with standard deviation. 
With the decoupling, $t_1$ and $t_2$ are more likely to appear after the derivation, since they are no longer strictly tied to the shared representations. More importantly, the performance is better with the decoupling. 
We see that the coupling effect harms the search quality for image classification but not much for object detection and instance segmentation. We think that it is because $t_1$ and $t_2$ are not as favored in those two tasks as in image classification, since the two tasks benefit more from the combination of larger RFs.

%% file: tables/cifar.tex
\begin{table}[t]
\centering
\footnotesize
\caption{{\bf Ablation on the decoupling.}
Test error (lower is better) on CIFAR-10. 
\emph{Shared Trans.} refers to the transformations that correspond to the shared representations (\ie $t_1$ and $t_2$  in Figure 2 in the main paper), and \emph{Other Trans.} denotes the $t_3$ to $t_5$. 
}
\vspace{-1mm}

\begin{tabular}{l@{~~}c@{~~~~}c@{~~~~}c@{~~~~}c@{~~~~}c}
\hline
\multirow{2}{*}{Method}  & \multirow{2}{*}{Decouple}  &  Shared Trans. &  Other Trans. & Params  & Test Error    \\ 
 & & (\%) & (\%) & (M) & (\%) \\ \hline

w/o \rep  &  -  &  87.1 & 12.9 & 3.3 & $2.93 \pm 0.11$ \\
\hline

\multirow{2}{*}{w \rep} & \xmark  & 30.0 & 70.0 & 3.8 & $3.15 \pm 0.12$ \\
 & \cmark & 85.7 & 14.3  & 3.2 & $2.98 \pm 0.08$  \\

\hline

\end{tabular}

\vspace{-1mm}
\label{tab:cifar}
\end{table}

%% file: main.bbl
\begin{thebibliography}{10}
\providecommand{\url}[1]{\texttt{#1}}
\providecommand{\urlprefix}{URL }
\providecommand{\doi}[1]{https://doi.org/#1}

\bibitem{chen2019detnas}
Chen, Y., Yang, T., Zhang, X., Meng, G., Pan, C., Sun, J.: Det{NAS}: Backbone
  search for object detection. arXiv preprint arXiv:1903.10979  (2019)

\bibitem{dai2016r}
Dai, J., Li, Y., He, K., Sun, J.: {R-FCN}: Object detection via region-based
  fully convolutional networks. In: Advances in neural information processing
  systems. pp. 379--387 (2016)

\bibitem{dai2017deformable}
Dai, J., Qi, H., Xiong, Y., Li, Y., Zhang, G., Hu, H., Wei, Y.: Deformable
  convolutional networks. In: Proceedings of the IEEE international conference
  on computer vision. pp. 764--773 (2017)

\bibitem{duan2019centernet}
Duan, K., Bai, S., Xie, L., Qi, H., Huang, Q., Tian, Q.: {CenterNet}: Keypoint
  triplets for object detection. In: Proceedings of the IEEE International
  Conference on Computer Vision. pp. 6569--6578 (2019)

\bibitem{everingham2010pascal}
Everingham, M., Van~Gool, L., Williams, C.K., Winn, J., Zisserman, A.: The
  pascal visual object classes ({VOC}) challenge. International journal of
  computer vision  \textbf{88}(2),  303--338 (2010)

\bibitem{ghiasi2018dropblock}
Ghiasi, G., Lin, T.Y., Le, Q.V.: {DropBlock}: A regularization method for
  convolutional networks. In: Advances in Neural Information Processing
  Systems. pp. 10727--10737 (2018)

\bibitem{ghiasi2019fpn}
Ghiasi, G., Lin, T.Y., Le, Q.V.: {NAS-FPN}: Learning scalable feature pyramid
  architecture for object detection. In: Proceedings of the IEEE Conference on
  Computer Vision and Pattern Recognition. pp. 7036--7045 (2019)

\bibitem{goldberg1991comparative}
Goldberg, D.E., Deb, K.: A comparative analysis of selection schemes used in
  genetic algorithms. In: Foundations of genetic algorithms, vol.~1, pp.
  69--93. Elsevier (1991)

\bibitem{guo2020hit}
Guo, J., Han, K., Wang, Y., Zhang, C., Yang, Z., Wu, H., Chen, X., Xu, C.:
  {Hit-Detector}: Hierarchical trinity architecture search for object
  detection. In: Proceedings of the IEEE/CVF Conference on Computer Vision and
  Pattern Recognition. pp. 11405--11414 (2020)

\bibitem{he2017mask}
He, K., Gkioxari, G., Doll{\'a}r, P., Girshick, R.: Mask {R-CNN}. In:
  Proceedings of the IEEE international conference on computer vision. pp.
  2961--2969 (2017)

\bibitem{he2016deep}
He, K., Zhang, X., Ren, S., Sun, J.: Deep residual learning for image
  recognition. In: Proceedings of the IEEE conference on computer vision and
  pattern recognition. pp. 770--778 (2016)

\bibitem{howard2017mobilenets}
Howard, A.G., Zhu, M., Chen, B., Kalenichenko, D., Wang, W., Weyand, T.,
  Andreetto, M., Adam, H.: Mobilenets: Efficient convolutional neural networks
  for mobile vision applications. arXiv preprint arXiv:1704.04861  (2017)

\bibitem{huang2019mask}
Huang, Z., Huang, L., Gong, Y., Huang, C., Wang, X.: Mask scoring {R-CNN}. In:
  Proceedings of the IEEE Conference on Computer Vision and Pattern
  Recognition. pp. 6409--6418 (2019)

\bibitem{krizhevsky2009learning}
Krizhevsky, A., Hinton, G., et~al.: Learning multiple layers of features from
  tiny images  (2009)

\bibitem{law2018cornernet}
Law, H., Deng, J.: {CornerNet}: Detecting objects as paired keypoints. In:
  Proceedings of the European Conference on Computer Vision (ECCV). pp.
  734--750 (2018)

\bibitem{li2019scale}
Li, Y., Chen, Y., Wang, N., Zhang, Z.: Scale-aware trident networks for object
  detection. arXiv preprint arXiv:1901.01892  (2019)

\bibitem{liang2019computation}
Liang, F., Lin, C., Guo, R., Sun, M., Wu, W., Yan, J., Ouyang, W.: Computation
  reallocation for object detection. arXiv preprint arXiv:1912.11234  (2019)

\bibitem{lin2017feature}
Lin, T.Y., Doll{\'a}r, P., Girshick, R., He, K., Hariharan, B., Belongie, S.:
  Feature pyramid networks for object detection. In: Proceedings of the IEEE
  conference on computer vision and pattern recognition. pp. 2117--2125 (2017)

\bibitem{lin2017focal}
Lin, T.Y., Goyal, P., Girshick, R., He, K., Doll{\'a}r, P.: Focal loss for
  dense object detection. In: Proceedings of the IEEE international conference
  on computer vision. pp. 2980--2988 (2017)

\bibitem{lin2014microsoft}
Lin, T.Y., Maire, M., Belongie, S., Hays, J., Perona, P., Ramanan, D.,
  Doll{\'a}r, P., Zitnick, C.L.: Microsoft {COCO}: Common objects in context.
  In: European conference on computer vision. pp. 740--755. Springer (2014)

\bibitem{liu2019auto}
Liu, C., Chen, L.C., Schroff, F., Adam, H., Hua, W., Yuille, A.L., Fei-Fei, L.:
  Auto-deeplab: Hierarchical neural architecture search for semantic image
  segmentation. In: Proceedings of the IEEE Conference on Computer Vision and
  Pattern Recognition. pp. 82--92 (2019)

\bibitem{liu2018progressive}
Liu, C., Zoph, B., Neumann, M., Shlens, J., Hua, W., Li, L.J., Fei-Fei, L.,
  Yuille, A., Huang, J., Murphy, K.: Progressive neural architecture search.
  In: Proceedings of the European Conference on Computer Vision (ECCV). pp.
  19--34 (2018)

\bibitem{liu2018darts}
Liu, H., Simonyan, K., Yang, Y.: {DARTS}: Differentiable architecture search.
  arXiv preprint arXiv:1806.09055  (2018)

\bibitem{liu2018receptive}
Liu, S., Huang, D., et~al.: Receptive field block net for accurate and fast
  object detection. In: Proceedings of the European Conference on Computer
  Vision (ECCV). pp. 385--400 (2018)

\bibitem{liu2016ssd}
Liu, W., Anguelov, D., Erhan, D., Szegedy, C., Reed, S., Fu, C.Y., Berg, A.C.:
  {SSD}: Single shot multibox detector. In: European conference on computer
  vision. pp. 21--37. Springer (2016)

\bibitem{luo2018neural}
Luo, R., Tian, F., Qin, T., Chen, E., Liu, T.Y.: Neural architecture
  optimization. In: Advances in neural information processing systems. pp.
  7816--7827 (2018)

\bibitem{peng2019efficient}
Peng, J., Sun, M., Zhang, Z., Tan, T., Yan, J.: Efficient neural architecture
  transformation searchin channel-level for object detection. arXiv preprint
  arXiv:1909.02293  (2019)

\bibitem{pham2018efficient}
Pham, H., Guan, M.Y., Zoph, B., Le, Q.V., Dean, J.: Efficient neural
  architecture search via parameter sharing. arXiv preprint arXiv:1802.03268
  (2018)

\bibitem{real2019regularized}
Real, E., Aggarwal, A., Huang, Y., Le, Q.V.: Regularized evolution for image
  classifier architecture search. In: Proceedings of the AAAI Conference on
  Artificial Intelligence. vol.~33, pp. 4780--4789 (2019)

\bibitem{real2017large}
Real, E., Moore, S., Selle, A., Saxena, S., Suematsu, Y.L., Tan, J., Le, Q.V.,
  Kurakin, A.: Large-scale evolution of image classifiers. In: Proceedings of
  the 34th International Conference on Machine Learning-Volume 70. pp.
  2902--2911. JMLR. org (2017)

\bibitem{redmon2018yolov3}
Redmon, J., Farhadi, A.: {YOLO}v3: An incremental improvement. arXiv preprint
  arXiv:1804.02767  (2018)

\bibitem{ren2015faster}
Ren, S., He, K., Girshick, R., Sun, J.: Faster {R-CNN}: Towards real-time
  object detection with region proposal networks. In: Advances in neural
  information processing systems. pp. 91--99 (2015)

\bibitem{sifre2014rigid}
Sifre, L., Mallat, S.: Rigid-motion scattering for image classification. Ph. D.
  dissertation  (2014)

\bibitem{simonyan2014very}
Simonyan, K., Zisserman, A.: Very deep convolutional networks for large-scale
  image recognition. arXiv preprint arXiv:1409.1556  (2014)

\bibitem{singh2018analysis}
Singh, B., Davis, L.S.: An analysis of scale invariance in object detection
  snip. In: Proceedings of the IEEE conference on computer vision and pattern
  recognition. pp. 3578--3587 (2018)

\bibitem{singh2018sniper}
Singh, B., Najibi, M., Davis, L.S.: {SNIPER}: Efficient multi-scale training.
  In: Advances in neural information processing systems. pp. 9310--9320 (2018)

\bibitem{stamoulis2019single}
Stamoulis, D., Ding, R., Wang, D., Lymberopoulos, D., Priyantha, B., Liu, J.,
  Marculescu, D.: {Single-Path NAS}: Designing hardware-efficient convnets in
  less than 4 hours. arXiv preprint arXiv:1904.02877  (2019)

\bibitem{szegedy2015going}
Szegedy, C., Liu, W., Jia, Y., Sermanet, P., Reed, S., Anguelov, D., Erhan, D.,
  Vanhoucke, V., Rabinovich, A.: Going deeper with convolutions. In:
  Proceedings of the IEEE conference on computer vision and pattern
  recognition. pp.~1--9 (2015)

\bibitem{tian2019fcos}
Tian, Z., Shen, C., Chen, H., He, T.: Fcos: Fully convolutional one-stage
  object detection. arXiv preprint arXiv:1904.01355  (2019)

\bibitem{wang2019fcos}
Wang, N., Gao, Y., Chen, H., Wang, P., Tian, Z., Shen, C.: {NAS-FCOS}: Fast
  neural architecture search for object detection. arXiv preprint
  arXiv:1906.04423  (2019)

\bibitem{wu2018group}
Wu, Y., He, K.: Group normalization. In: Proceedings of the European Conference
  on Computer Vision (ECCV). pp. 3--19 (2018)

\bibitem{xie2017aggregated}
Xie, S., Girshick, R., Dollár, P., Tu, Z., He, K.: Aggregated residual
  transformations for deep neural networks  (2017)

\bibitem{xu2019auto}
Xu, H., Yao, L., Zhang, W., Liang, X., Li, Z.: {Auto-FPN}: Automatic network
  architecture adaptation for object detection beyond classification. In:
  Proceedings of the IEEE International Conference on Computer Vision. pp.
  6649--6658 (2019)

\bibitem{yao2019sm}
Yao, L., Xu, H., Zhang, W., Liang, X., Li, Z.: {SM-NAS}: Structural-to-modular
  neural architecture search for object detection. arXiv preprint
  arXiv:1911.09929  (2019)

\bibitem{zhang2020bridging}
Zhang, S., Chi, C., Yao, Y., Lei, Z., Li, S.Z.: Bridging the gap between
  anchor-based and anchor-free detection via adaptive training sample
  selection. In: Proceedings of the IEEE/CVF Conference on Computer Vision and
  Pattern Recognition. pp. 9759--9768 (2020)

\bibitem{zhang2019freeanchor}
Zhang, X., Wan, F., Liu, C., Ji, R., Ye, Q.: {FreeAnchor}: Learning to match
  anchors for visual object detection. In: Advances in Neural Information
  Processing Systems. pp. 147--155 (2019)

\bibitem{zhu2019soft}
Zhu, C., Chen, F., Shen, Z., Savvides, M.: Soft anchor-point object detection.
  arXiv preprint arXiv:1911.12448  (2019)

\bibitem{zhu2019feature}
Zhu, C., He, Y., Savvides, M.: Feature selective anchor-free module for
  single-shot object detection. arXiv preprint arXiv:1903.00621  (2019)

\bibitem{zoph2016neural}
Zoph, B., Le, Q.V.: Neural architecture search with reinforcement learning.
  arXiv preprint arXiv:1611.01578  (2016)

\end{thebibliography}
